# Almost-everywhere algorithmic stability and generalization error

Samuel Kutin* and Partha Niyogi[†]

## Abstract

We explore in some detail the notion of algorithmic stability as a viable framework for analyzing the generalization error of learning algorithms. We introduce the new notion of training stability of a learning algorithm and show that, in a general setting, it is sufficient for good bounds on generalization error. In the PAC setting, training stability is both necessary and sufficient for learnability.

The approach based on training stability makes no reference to VC dimension or VC entropy. There is no need to prove uniform convergence, and generalization error is bounded directly via an extended McDiarmid inequality. As a result it potentially allows us to deal with a broader class of learning algorithms than Empirical Risk Minimization.

We also explore the relationships among VC dimension, generalization error, and various notions of stability. Several examples of learning algorithms are considered.

## 1 Introduction

A major concern for learning theory is to develop frameworks to analyze the generalization error of learning algorithms. To date, the dominant theoretical paradigm for this has been the theory of Vapnik and Chervonenkis. Notions of compactness, VC dimension, or $V_\gamma$ dimension are used to prove uniform convergence from which bounds on generalization error follow as a natural consequence.

* Department of Computer Science, University of Chicago, 100 E. 58th Street, Chicago, IL 60637. Email: kutin@cs.uchicago.edu.
† Department of Computer Science, University of Chicago, 100 E. 58th Street, Chicago, IL 60637. Email: niyogi@cs.uchicago.edu.

In this paper, we consider the notion of algorithmic stability and show that it provides a viable alternative framework within which generalization error of learning algorithms may be analyzed. We consider several different types of algorithmic stability and introduce, in particular, the new notion of *training stability*. We show that it is sufficient to provide exponential bounds on generalization error and in the PAC setting it is both necessary and sufficient for this purpose.

Algorithmic stability was first introduced by Devroye and Wagner [3]. An algorithm is stable at a training set $S$ if any change of a single point in $S$ yields only a small change in the output hypothesis. (We refer to this notion as "weak hypothesis stability.") Breiman [2] recognizes the importance of stability when he argues that unstable weak learners benefit from randomization algorithms such as bagging. Kearns and Ron [5] consider the weaker, related notion of error stability. They prove bounds on the error of leave-one-out estimates of error rates and not on the generalization error directly. Furthermore, their arguments rely on the traditional notion of VC dimension [10].

Recently, Bousquet and Elisseeff [1] proved that algorithms that are "uniformly hypothesis stable" (i.e., stable at every training set) have low generalization error; their proof does not make any reference to VC dimension. By avoiding VC theory, stability allows us to focus on a wider class of learning algorithms than empirical risk minimization (ERM). For example, they show that regularization networks are stable [1]. While their result constitutes an important first step, the notion of uniform hypothesis stability is too restrictive, and several natural algorithms violate this condition.

In Section 3, we discuss several notions of almost-everywhere stability. Through examples, we argue that some notions are too restrictive, while others are too broad. We propose a new notion of stability, *training stability*: most of the time, changing one point in a training set leads to a small change in the error on points in that set.

Our main result is Theorem 4.4: training stability implies good concentration bounds on generalization error. Our proof, which appears in Section 4, follows the argument of Bousquet and Elisseeff [1]. They use McDiarmid's method of independent bounded differences [9]. We use an extension of McDiarmid's Theorem [6], which we state in Section 2.1.

In Section 5, we show that, in the PAC setting, training stability is necessary and sufficient for PAC learnability. We give examples of stable algorithms in Section 6. We discuss some open questions in Section 7.

**Note 1.1** In the long version of this note [8], the authors define additional notions of stability, and discuss examples in greater depth. This version contains the central definitions and results.

## 2 Preliminaries

There is a space $X$ of points, or *instances*, with some unknown distribution $\Delta$. A target operator takes as input an element $x \in X$ and outputs some $y$ in a set $Y$ of labels, according to a conditional distribution function $F(y \mid x)$; often $F$ is given by a deterministic target function $f: X \to Y$. For simplicity, we assume $Y = \{-1, 1\}$. We let $Z = X \times Y$, the space of *examples*. We write $D$ for the distribution on $Z$ induced by $\Delta$ and $F$.

**Note 2.1** If $D$ is a distribution on a set $Z$, we use the notation $z \sim D$ to mean that $z$ is chosen from $Z$ according to distribution $D$.

**Note 2.2** If we have some probability space $\Omega$, and some property $\Phi: \Omega \to \{\text{true}, \text{false}\}$, then the notation

$$\forall^\delta \omega, \quad \Phi(\omega)$$

means that $\Pr_{\omega \in \Omega}(\Phi(\omega)) \geq 1 - \delta$. In other words, property $\Phi$ holds for all but a $\delta$ fraction of $\Omega$.

A *classifier* or *hypothesis* is a function $h: X \to [-1, 1]$. (A value of $h(x)$ in $(-1, 1)$ corresponds to a confidence-rated prediction.) We use $\mathcal{H}$ to denote a space of classifiers.

**Definition 2.3** The *cost* of a classifier $h$ on a point $z \in Z$ is a measure of the error $h$ makes on $z$. We denote this cost by $c(h, z)$, and we require $0 \leq c(h, z) \leq M$ for some constant $M$.

**Definition 2.4** The error rate of a classifier $h: X \to [-1, 1]$, with respect to a distribution $D$, is

$$\text{Err}_D(h) = \mathbf{E}_{z \sim D}(c(h, z)).$$

For $S \in Z^m$, we use the notation $\text{Err}_S$ to mean $\text{Err}_p$ where $p$ is the uniform distribution on $S$.

**Definition 2.5** A *learning algorithm* $\mathcal{A}$ is a process which takes as input a finite training set $S \in \bigcup_m Z^m$ and outputs a function $f_S: X \to [-1, 1]$.

Informally, a learning algorithm is given a finite subset of $Z$, and attempts to construct a classifier $h$ such that $h(x)$ is a good approximation to the output of the target operator on input $x$. For simplicity, we assume that all of our learning algorithms are symmetric.

**Note 2.6** For $S \in Z^m$, $S = (z_1, \ldots, z_m)$, we let $S^i$ denote $S \setminus z_i$, the training set with the $i$th instance removed. So, for each $i$, $S^i \in Z^{m-1}$. For $u \in Z$, we let $S^{i,u}$ denote $S^i \cup \{u\}$, the training set with the $i$th instance replaced by $u$. So $S^{i,u} \in Z^m$.

**Definition 2.7** The *training error* of a learning algorithm on an input $S$ is the error rate of $f_S$ on the set $S$, or $\text{Err}_S(f_S)$. The *true error* of a learning algorithm on an input $S$ is the error rate of $f_S$ on a randomly chosen example, or $\text{Err}_D(f_S)$. The *generalization error* of a learning algorithm on an input $S$ is the difference between the observed error rate and the true error rate:

$$\text{gen}(S) = \text{Err}_D(f_S) - \text{Err}_S(f_S).$$

We write $\mu = \mathbf{E}_S(\text{gen}(S))$. Hence $\mu$ depends on $m$.

Our focus is on bounding the probability that the generalization error of a learning algorithm $\mathcal{A}$ is large. Note that this is not sufficient to imply that $\mathcal{A}$ is a good algorithm. However, it is easier to determine empirically whether an algorithm has good training error. Also, in general, generalization error bounds are more elusive than training error bounds.

### 2.1 Extensions of McDiarmid's Inequality

McDiarmid's method of independent bounded differences [9] gives concentration bounds on multivariate functions in terms of the maximum effect that changing one coordinate of the input can have on the output. We will use a generalization [6] of McDiarmid's Theorem which applies when changing one coordinate usually leads to a small change in the output, but not always. We need this result to extend Bousquet and Elisseeff's argument [1] that uniform hypothesis stability gives good bounds on generalization error to weaker notions of stability.

**Definition 2.8 ([6])** Let $\Omega_1, \ldots, \Omega_m$ be probability spaces. Let $\Omega = \prod_{k=1}^m \Omega_k$, and let $X$ be a random variable on $\Omega$. We say that $X$ is *weakly difference-bounded* by $(b, c, \delta)$ if the following holds: for any $k$,

$$\forall^\delta (\omega, \upsilon) \in \Omega \times \Omega_k, \quad |X(\omega) - X(\omega')| \leq c,$$

where $\omega'_k = \upsilon$ and $\omega'_i = \omega_i$ for $i \neq k$. In words, if we choose $\omega \in \Omega$, and $\upsilon \in \Omega_k$, and we construct $\omega'$ by replacing the $k$th entry of $\omega$ with $\upsilon$, then the inequality holds for all but a $\delta$ fraction of the choices. Furthermore, for any $\omega$ and $\omega'$ differing only in the $k$th coordinate,

$$|X(\omega) - X(\omega')| \leq b.$$

**Theorem 2.9 ([6])** *Let $\Omega_1, \ldots, \Omega_m$ be probability spaces. Let $\Omega = \prod_{k=1}^m \Omega_k$, and let $X$ be a random variable on $\Omega$ which is weakly difference-bounded by $(b, \frac{\lambda}{m}, \exp(-Km))$. Let $\mu = \mathbf{E}(X)$. If $0 < \tau \leq T(b,\lambda,K)$, and $m \geq N(b,\lambda,K,\tau)$, then*

$$\Pr(|X - \mu| \geq \tau) \leq 4\exp\left(-\frac{\tau^2 m}{40\lambda^2}\right).$$

The bounds $T$ and $N$ above, and the requirement that $c = \frac{\lambda}{m}$ and $\delta = e^{-Km}$, are necessary only for the simplified version of the theorem. More general versions [6] apply for other choices of the parameters. The above version applies when $\tau = \Theta(1/\sqrt{m})$, which is the case of interest.

**Fact 2.10** *Let $X, Y$ be random variables on $\Omega = \prod_{k=1}^m \Omega_k$. If $X$ is weakly difference-bounded by $(b_1, c_1, \delta_1)$, and $Y$ is weakly difference-bounded by $(b_2, c_2, \delta_2)$, then $X + Y$ is weakly difference-bounded by $(b_1 + b_2, c_1 + c_2, \delta_1 + \delta_2)$.*

## 3 Notions of stability

### 3.1 Uniform hypothesis stability

**Definition 3.1 (Bousquet and Elisseeff [1])** A learning algorithm $\mathcal{A}$ is *uniformly $\beta$-hypothesis-stable*, or has *uniform hypothesis stability* $\beta$, if the following holds:

$$\forall S \in Z^m, \forall i, \forall u, z \in Z, \quad |c(f_S, z) - c(f_{S^{i,u}}, z)| \leq \beta.$$

We view $\beta$ as a function of $m$. We are most interested in the case where $\beta = \lambda/m$ for a constant $\lambda$. Bousquet and Elisseeff [1] prove that regularization has uniform hypothesis stability $O(1/m)$.

Bousquet and Elisseeff [1] were the first to obtain bounds on generalization error:

**Theorem 3.2 (Bousquet and Elisseeff [1])** *If $\mathcal{A}$ has uniform hypothesis stability $\beta$, then, for all $\tau > 0$,*

$$\Pr(|\operatorname{gen}(S)| > \tau + \beta) \leq 2\exp\left(\frac{-\tau^2 m}{2(m\beta + M)^2}\right).$$

Theorem 3.2 gives good bounds on generalization error when $\beta = O(1/m)$. Our goal in this paper is to extend Bousquet and Elisseeff's Theorem 3.2 to weaker notions of stability. We first motivate this analysis by presenting some simple learning algorithms which are not uniformly hypothesis-stable.

**Example 3.3** Suppose that $\mathcal{A}$ is a $\pm 1$-*algorithm*: for all $S \in Z^m$, and for all $x \in X$, $f_S(x) \in \{-1, 1\}$. If $\mathcal{A}$ is uniformly $\beta$-hypothesis-stable for some $\beta < 1$, then $\mathcal{A}$ must be the constant algorithm: there is some $h$ where $f_S = h$ for every training set $S$.

We should note that Bousquet and Elisseeff [1] discuss the setting in which the binary value $f_S(x)$ is obtained by thresholding a real-valued quantity. In this case, uniform hypothesis stability can be applied.

**Example 3.4** Let $\mathcal{H}$ be a finite collection of classifiers, and let $\mathcal{A}$ be a learning algorithm which performs ERM over $\mathcal{H}$. Then, assuming $\mathcal{A}$ is not constant, $\mathcal{A}$ is not uniformly $\beta$-hypothesis-stable for any $\beta = o(1)$.

We see this by dividing $Z^m$ into regions: for each $h \in \mathcal{H}$, $R(h) = \{S \in Z^m \mid f_S = h\}$. Some training sets $S$ must lie on the boundary between regions, so the stability $\beta$ is at least

$$\min_{h, h' \in \mathcal{H}} \left\{ \sup_{z \in Z}\{|c(h, z) - c(h', z)|\} \right\}.$$

### 3.2 Weak hypothesis stability

We now introduce a weaker notion of stability, *weak hypothesis stability*, which allows us to handle Examples 3.3 and 3.4.

**Definition 3.5** A learning algorithm $\mathcal{A}$ is *weakly $(\beta, \delta)$-hypothesis-stable*, or has *weak hypothesis stability* $(\beta, \delta)$, if, for any $i \in \{1, \ldots, m\}$,

$$\forall^{\delta} S, u, \quad \max_{z \in Z}\{|c(f_S, z) - c(f_{S^{i,u}}, z)|\} \leq \beta,$$

where $S \sim D^m$ and $u \sim D$.

**Remark 3.6** Stability is implicit in the work of Devroye and Wagner [3]. Kearns and Ron formulate a related definition, which they call hypothesis stability but which we would call *weak $L_1$ stability* [8]. Bousquet and Elisseeff [1] use the term hypothesis stability for a slightly different definition, phrased in terms of expectations.

The parameters $\beta$ and $\delta$ are functions of $m$. We will chiefly be interested in the case where $\beta = O(1/m)$ and $\delta = \exp(-\Omega(m))$. In some natural examples, we have $\beta = 0$.

We will show in Section 4 that weak hypothesis stability is sufficient for good bounds on generalization error. However, even weak hypothesis stability is too restrictive for our purposes. It is possible for ERM over a space of VC dimension 1 not to be weakly hypothesis stable.

**Example 3.7** Let $X = [0, 1]$. We consider the class $\mathcal{H}$ of threshold functions $g_\theta(x) = \text{sign}(x - \theta)$. (We assume for purposes of discussion that $g_\theta(\theta) = 1$.)

Let $\mathcal{A}$ be an algorithm performing ERM over this space $\mathcal{H}$ as follows: let $a$ be the maximal value for which $(a, -1) \in S$, and let $b$ be the minimal value for which $(b, 1) \in S$. Assuming the data is labeled according to some $g_\theta$, we must have $\theta \in (a, b]$.

Let $\xi = \frac{1}{m} \sum_{i=1}^{m} x_i$, the average value of a point in $S$. We let $f_S = g_{\theta_S}$, where $\theta_S = a + \xi(b - a)$.

Since $\theta_S \in (a, b]$, the training error $\text{Err}_S(f_S) = 0$. Hence, we can say that $\mathcal{A}$ performs ERM. Note that $\text{VCdim}(\mathcal{H}) = 1$.

Any small change to $S$ will yield a small change to $\theta_S$, so $f_S$ and $f_{S'}$ differ somewhere. This is enough to force $\max_z |c(f_S, z) - c(f_{S'}, z)|$ to be 1.

Example 3.7 demonstrates that ERM over a space of finite VC dimension is not necessarily weakly hypothesis stable. Of course, it is possible to construct another algorithm performing ERM over the same space which is weakly hypothesis stable. We return to this idea in Example 6.6.

In Section 3.4, we will introduce the notion of training stability. We will see that the algorithm of Example 3.7 is training stable.

### 3.3 Weak error stability

We now discuss an even weaker notion of stability.

**Definition 3.8 (Kearns and Ron [5])** A learning algorithm $\mathcal{A}$ is *weakly $(\beta, \delta)$-error-stable*, or has *weak error stability* $(\beta, \delta)$, if, for any $i \in \{1, \ldots, m\}$,

$$\forall^\delta S, u, \quad |\text{Err}_D(f_S) - \text{Err}_D(f_{S^{i,u}})| \leq \beta,$$

where $S \sim D^m$ and $u \sim D$.

It is clear that, for any $S$ and $S'$, $|\text{Err}_D(f_S) - \text{Err}_D(f_{S'})| \leq M$. This implies the following:

**Fact 3.9** *If $\mathcal{A}$ is weakly $(\beta, \delta)$-error-stable, then $\text{Err}_D(f_S)$ is weakly difference-bounded by $(M, \beta, \delta)$.*

**Remark 3.10** Kearns and Ron [5] refer to this notion simply as error stability. They prove that ERM over a space of finite VC dimension is weakly error stable. They also prove bounds on the error of leave-one-out estimates using weak error stability, but their arguments require an assumption of finite VC dimension.

The notion of weak error stability is general enough to be our primary definition of stability; in particular, it applies to ERM over a space of finite VC dimension. However, weak error stability is not strong enough to imply good bounds on generalization error.

**Example 3.11** Let $\mathcal{X} = [0, 1]$. We have some function $d : \mathbb{N} \to (0, 1]$, where $d(m) = o(1/m)$. Given a training set $S = \{(x_i, y_i)\}$, we define $f_S$ to be the following function from $\mathcal{X}$ to $\{-1, 1\}$:

1. Given $x \in \mathcal{X}$, let $j = \text{argmin}_{1 \leq i \leq m} |x - x_i|$. So $x_j$ is the nearest neighbor to $x$ in $\mathcal{X}$.

2. If $|x - x_j| < d(m)$, return $y_j$; otherwise, return 1.

This defines a learning algorithm $\mathcal{A}: S \mapsto f_S$.

If we take $d(m) = 0$, then $\mathcal{A}$ returns the correct labels on the training set and 1 elsewhere. For $d(m) > 0$, we use a nearest-neighbor approximation near the training points.

Given any two training sets $S$ and $S'$ which differ in one element, $f_S$ and $f_{S'}$ differ on a region of size at most $4d(m)$. So $\mathcal{A}$ has weak error stability $(4d(m), 0)$.

However, let $\eta$ be the error rate (with respect to $D$) of the constant hypothesis 1. Since the measure of $\{x : f_S(x) = -1\}$ is at most $2md(m)$, the generalization error is at least

$$\text{gen}(S) = \text{Err}_D(f_S) \geq \eta - 2md(m).$$

Since $d(m) = o(1/m)$, we have $\text{gen}(S) \to \eta$ as $m \to \infty$.

Hence, weak error stability is not sufficient to prove bounds on generalization error.

### 3.4 Training stability and CV stability

We observe in Example 3.11 that weak error stability is not sufficient to prove good bounds on generalization error. Weak error stability implies that $\text{Err}_D(f_S)$ is concentrated about its mean. As we will see in Section 4, there are two other statements we must prove for our main result: the mean generalization error $\mu$ is small, and the training error $\text{Err}_S(f_S)$ is concentrated about its mean.

In this section, we introduce new notions of stability which are chosen precisely to complete the proof described above. We first introduce *CV stability*, which will give us a bound on $\mu$. We then introduce *overlap stability*, which we will use to show that training error is concentrated about its mean.

We show in Section 3.4.1 that CV stability implies weak error stability. Hence, we conclude Theorem 4.4: the combination of CV stability and overlap stability, which we call *training stability*, is sufficient for good bounds on generalization error.

**Definition 3.12** A learning algorithm $\mathcal{A}$ is $(\beta, \delta)$-*cross-validation-stable*, or $(\beta, \delta)$-*CV-stable*, if, for any $i \in \{1, \ldots, m\}$,

$$\forall^\delta S, u \quad |c(f_S, u) - c(f_{S^{i,u}}, u)| \leq \beta.$$

We also say that $\mathcal{A}$ has *cross-validation stability* or *CV stability* $(\beta, \delta)$.

**Definition 3.13** A learning algorithm $\mathcal{A}$ is $(\beta, \delta)$-*overlap-stable*, or has *overlap stability* $(\beta, \delta)$, if, for any $i \in \{1, \ldots, m\}$,

$$\forall^\delta S, u, \quad |\operatorname{Err}_{S^i}(f_S) - \operatorname{Err}_{S^i}(f_{S^{i,u}})| \leq \beta.$$

We call this notion overlap stability because it says that, for most training sets $S$, $S'$ differing in only one coordinate, $f_S$ and $f_{S'}$ have similar performance on $S \cap S'$. (Recall that $S^i$ is $S$ with the $i$th example removed, so $S^i = S \cap S'$.)

We now combine these two notions into one definition. Since both apply to the performance of $S$ and $S'$ on the training set $S \cup S'$, we call this joint notion training stability:

**Definition 3.14** A learning algorithm $\mathcal{A}$ is $(\beta, \delta)$-*training-stable*, or has *training stability* $(\beta, \delta)$, if (1) $\mathcal{A}$ has CV stability $(\beta, \delta)$, and (2) $\mathcal{A}$ has overlap stability $(\beta, \delta)$.

**Observation 3.15** *Weak hypothesis stability* $(\beta, \delta)$ *implies training stability* $(\beta, \delta)$.

### 3.4.1 CV stability implies weak error stability

As we remark in Observation 3.15, weak hypothesis stability implies training stability (and, therefore, CV stability). We now show that CV stability implies weak error stability.

**Theorem 3.16** *Let $\mathcal{A}$ be a $(\beta, \delta)$-CV-stable learning algorithm. Then, for any function $\alpha(m) > 0$, $\mathcal{A}$ has weak error stability $(2\beta + 2M\alpha, 2\delta/\alpha)$.*

**Proof:** Fix $i$. Let $B$ be the set of "bad" $S$ for which

$$\Pr_z(|c(f_S, z) - c(f_{S^{i,z}}, z)| > \beta) > \alpha.$$

By the definition of CV stability,

$$\delta \geq \Pr_{S,z}(|c(f_S, z) - c(f_{S^{i,z}}, z)| > \beta)$$
$$= \Pr_{S,z}(|c(f_S, z) - c(f_{S^{i,z}}, z)| > \beta \mid S \in B) \Pr_S(S \in B)$$
$$\geq \alpha \Pr_S(S \in B),$$

so $\Pr_S(S \in B) \leq \delta/\alpha$.

Similarly, $\Pr_{S,u}(S^{i,u} \in B) \leq \delta/\alpha$. So, for all but a $2\delta/\alpha$ fraction of choices of $S, u$, both $S$ and $S^{i,u}$ are good.

Now, suppose that $S$ and $S^{i,u}$ are good. Then

$$\forall^\alpha z \quad |c(f_S, z) - c(f_{S^{i,z}}, z)| \leq \beta. \quad (1)$$
$$\forall^\alpha z \quad |c(f_{S^{i,u}}, z) - c(f_{S^{i,z}}, z)| \leq \beta. \quad (2)$$

(We use the fact that $(S^{i,u})^{i,z} = S^{i,z}$.)

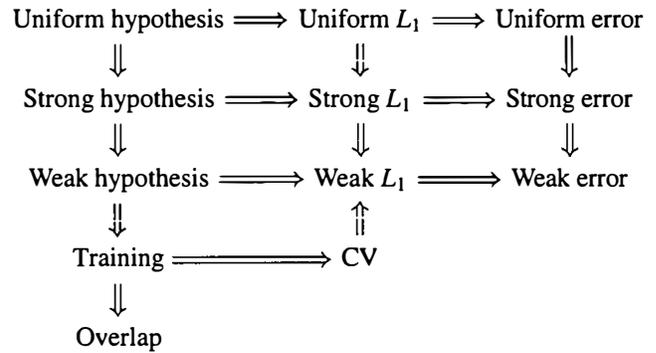

Figure 1: Implications among twelve notions of almost-everywhere stability

So, all but a $2\alpha$ fraction of $z$ satisfy (1) and (2). Hence, by the triangle inequality,

$$\forall^{2\alpha} z \quad |c(f_S, z) - c(f_{S^{i,u}}, z)| \leq 2\beta,$$

and therefore, whenever $S$ and $S^{i,u}$ are both good,

$$\mathbf{E}_z(|c(f_S, z) - c(f_{S^{i,u}}, z)|) \leq 2\beta + 2\alpha M.$$

Since $S$ and $S^{i,u}$ are both good with probability at least $1 - 2\delta/\alpha$, this implies weak error stability $(2\beta + 2\alpha M, 2\delta/\alpha)$. ■

We summarize the relationships between these different notions of stability in Figure 1. The figure includes other notions of stability discussed in detail in the extended version of this note [8].

## 4 Stability and generalization error

In this section, we prove Theorem 4.4: training stability implies good bounds on generalization error.

The proof proceeds in three main parts. Recall that $\operatorname{gen}(S) = \operatorname{Err}_D(f_S) - \operatorname{Err}_S(f_S)$, and that $\mu = \mathbf{E}(\operatorname{gen}(S))$.

1. Lemma 4.2: If $\mathcal{A}$ has CV stability $(\beta, \delta)$, then $|\mu| \leq \beta + \delta M$.

2. Lemma 4.3: If $\mathcal{A}$ has overlap stability $(\beta, \delta)$, then $\operatorname{Err}_S(f_S)$ is weakly difference-bounded by $(M, \beta + \frac{M}{m}, \delta)$.

3. Fact 3.9: If $\mathcal{A}$ has weak error stability $(\beta, \delta)$, then $\operatorname{Err}_D(f_S)$ is weakly difference-bounded by $(M, \beta, \delta)$.

Finally, we conclude that $\operatorname{gen}(S)$ is weakly difference-bounded and apply Theorem 2.9.

We begin by proving that CV stability implies that $\mu$ is small. We first state a technical lemma due to Bousquet and Elisseeff [1]:

**Lemma 4.1 (Bousquet and Elisseeff [1])** *For any learning algorithm $\mathcal{A}$, for any $i \in \{1, \ldots, m\}$,*

$$\mu = \mathbf{E}_{S,z \sim D^{m+1}} \left( c(f_S, z) - c(f_{S^{i,z}}, z) \right).$$

**Lemma 4.2** *If $\mathcal{A}$ is $(\beta, \delta)$ CV stable, $|\mu| \leq \beta + \delta M$.*

**Proof:** Fix some $i$. Let $\psi(S, u) = c(f_S, u) - c(f_{S^{i,u}}, u)$. By Lemma 4.1, $\mu = \mathbf{E}_{S, u \sim D^{m+1}}(\psi(S, u))$.

By CV stability, $|\psi(S, u)| \leq \beta$ with probability at least $1 - \delta$. For any $S$ and $u$, $|\psi(S, u)| \leq M$. We conclude that

$$\mathbf{E}_{S, u \sim D^{m+1}}(|\psi(S, u)|) \leq (1-\delta)\beta + \delta M \leq \beta + \delta M.$$

∎

We now consider the random variable $\mathrm{Err}_S(f_S)$.

**Lemma 4.3** *If $\mathcal{A}$ is $(\beta, \delta)$-overlap-stable, then $\mathrm{Err}_S(f_S)$ is weakly difference-bounded by $(M, \beta + \frac{M}{m}, \delta)$.*

**Proof:** Fix some $i \in \{1, \ldots, m\}$. Choose some $S = (z_1, \ldots, z_m) \sim D^m$ and $u \sim D$.

We have

$$\mathrm{Err}_S(f_S) = \frac{1}{m} c(f_S, z_i) + \frac{m-1}{m} \mathrm{Err}_{S^i}(f_S),$$

and similarly for $\mathrm{Err}_{S^{i,u}}(f_{S^{i,u}})$. So,

$$|\mathrm{Err}_S(f_S) - \mathrm{Err}_{S^{i,u}}(f_{S^{i,u}})| \leq$$
$$\frac{M}{m} + \frac{m-1}{m} |\mathrm{Err}_{S^i}(f_S) - \mathrm{Err}_{S^i}(f_{S^{i,u}})|.$$

Hence, by overlap stability,

$$\forall^\delta S, u, \quad |\mathrm{Err}_S(f_S) - \mathrm{Err}_{S^{i,u}}(f_{S^{i,u}})| \leq \frac{M}{m} + \frac{m-1}{m} \beta.$$

Trivially, $|\mathrm{Err}_S(f_S) - \mathrm{Err}_{S^{i,u}}(f_{S^{i,u}})| \leq M$ for every $S$ and $u$. This completes the proof. ∎

We are now ready to prove the main theorem.

**Theorem 4.4** *Suppose $\mathcal{A}$ is $\left(\frac{\lambda}{m}, e^{-Km}\right)$-training-stable. Then, for sufficiently large $m$, if $T_{\min} \leq \tau \leq T_{\max}$, we have*

$$\Pr(|\mathrm{gen}(S)| > \tau) \leq 4 \exp\left(\frac{-\tau^2 m}{1440(\lambda + M)^2}\right).$$

**Proof:** Let $\beta = \frac{\lambda}{m}$, and $\delta = e^{-Km}$.

Training stability implies overlap stability, so, by Lemma 4.3, $\mathrm{Err}_S(f_S)$ is weakly difference-bounded by $(M, \beta + \frac{M}{m}, \delta)$.

Also, training stability $(\beta, \delta)$ implies CV stability $(\beta, \delta)$, so, by Theorem 3.16 and Fact 3.9, $\mathrm{Err}_D(f_S)$ is weakly difference-bounded by $(M, 2\beta + \frac{2M}{m}, 2m\delta)$.

We conclude, using Fact 2.10, that $\mathrm{gen}(S)$ is weakly difference-bounded by $(2M, 3\beta + \frac{3M}{m}, (2m+1)\delta)$.

We now apply Theorem 2.9. For $\varepsilon \geq T$, and $m \geq N$,

$$\Pr(|\mathrm{gen}(S) - \mu| \geq \varepsilon) \leq 4 \exp\left(-\frac{\varepsilon^2 m}{360(\lambda + M)^2}\right).$$

By Lemma 4.2, $\mu \leq \beta + \delta M$. If $\varepsilon > \mu$, we write $\tau = 2\varepsilon$, proving the theorem. ∎

It turns out that $T_{\max}$ is a constant and $T_{\min} = O(1/m)$. So, for $\tau = \Theta(1/\sqrt{m})$, the theorem applies.

### 4.1 Other stability theorems

We can prove bounds on generalization error starting from other notions of stability. The proofs follow the same structure as that of Theorem 4.4; we first bound $|\mu|$, and then prove that $\mathrm{Err}_D(f_S)$ and $\mathrm{Err}_S(f_S)$ are weakly difference-bounded. We state two such results below, without proof.

**Theorem 4.5** *Suppose $\mathcal{A}$ is weakly $\left(\frac{\lambda}{m}, e^{-Km}\right)$-hypothesis-stable. Then, for sufficiently large $m$, if $T_{\min} \leq \tau \leq T_{\max}$, we have*

$$\Pr(|\mathrm{gen}(S)| > \tau) \leq 4 \exp\left(\frac{-\tau^2 m}{160(2\lambda + M)^2}\right).$$

**Theorem 4.6** *Suppose $\mathcal{A}$ performs ERM and has CV stability $\left(\frac{\lambda}{m}, e^{-Km}\right)$. Then, for sufficiently large $m$, if $T_{\min} \leq \tau \leq T_{\max}$, we have*

$$\Pr(|\mathrm{gen}(S)| > \tau) \leq 4 \exp\left(\frac{-\tau^2 m}{160(2\lambda + 3M)^2}\right).$$

In both theorems, $T_{\min}$ and $T_{\max}$ are of the same order as in Theorem 4.4.

## 5 CV stability and learnability

In this section, we discuss the relationship between CV stability and learnability.

Lemma 4.2 states that, if $\mathcal{A}$ is CV stable, then the average generalization error $\mu$ approaches 0 as $m \to \infty$. This implies that CV stable algorithms are good in the following sense: the expected error rate of the output classifier approaches the optimal error rate.

**Corollary 5.1** *Let $\mathcal{H}$ be a space of classifiers, and let $\mathcal{A}$ be a learning algorithm performing ERM over $\mathcal{H}$. Let $h_* \in \mathcal{H}$ be an optimal classifier. If $\mathcal{A}$ has CV stability $(\beta, \delta)$, where $\beta, \delta \to 0$ as $m \to \infty$, then $\mathbf{E}_S(\mathrm{Err}_D(f_S)) \to \mathrm{Err}_D(h_*)$ as $m \to \infty$.*

Note that this statement does not refer to the VC dimension of $\mathcal{H}$.

**Proof:** By Lemma 4.2,
$$\lim_{m\to\infty}(\mathbf{E}_S(\mathrm{Err}_D(f_S)) - \mathbf{E}_S(\mathrm{Err}_S(f_S))) = 0.$$

Let $h_* \in \mathcal{H}$ be an optimal classifier, so $\mathrm{Err}_D(h_*) \leq \mathrm{Err}_D(f_S)$. Clearly $\mathrm{Err}_D(h_*) = \mathbf{E}_S(\mathrm{Err}_S(h_*))$. Also, $\mathrm{Err}_S(f_S) \leq \mathrm{Err}_S(h_*)$. Combining these inequalities yields the desired result. ∎

In the PAC setting, we can say something stronger:

**Theorem 5.2** *Let $\mathcal{H}$ be a space of $\pm 1$-classifiers, and let $\mathcal{A}$ be a learning algorithm performing ERM over $\mathcal{H}$. Suppose that our examples are generated to be consistent with some $h_0 \in \mathcal{H}$.*

*Then $\mathcal{A}$ has CV stability $(0,\delta)$, with $\delta = \exp(-\Omega(m))$, if and only if $\mathrm{Err}_D(f_S) \to 0$ exponentially in $m$.*

**Remark 5.3** In the setting of Theorem 5.2, ERM always has overlap stability $(0,0)$. So CV stability is equivalent to training stability.

**Proof:** For $\pm 1$-classifiers, any $\beta < 1$ is equivalent; we assume $\beta = 0$. We also assume $M = 1$.

Since $\mathrm{Err}_D(h_0) = 0$, we have $\mathrm{Err}_S(h_0) = 0$ for all $S$. By the definition of ERM, we thus have $\mathrm{Err}_S(f_S) = 0$ for all $S$, so $\mathrm{gen}(S) = \mathrm{Err}_D(f_S)$. Also, for all $S$ and $z$, $c(\mathrm{Err}_{S^{i;z}}, z) = 0$. Hence,

$$\begin{aligned}
\text{CV stability } (0,\delta) &\iff \forall^\delta S, z \quad c(f_S, z) = c(f_{S^{i;z}}, z) \\
&\iff \mathbf{E}_{S,z}(c(f_S, z)) \leq \delta \\
&\iff \mathbf{E}_S(\mathrm{Err}_D(f_S)) \leq \delta.
\end{aligned}$$

So, $\mathbf{E}_S(\mathrm{Err}_D(f_S)) \to 0$ exponentially in $m$ if and only if $\mathcal{A}$ is $(0,\delta)$-CV-stable where $\delta \to 0$ exponentially in $m$. By Theorem 4.4, CV stability with such a $\delta$ implies exponential concentration bounds on $\mathrm{gen}(S)$, and hence on $\mathrm{Err}_D(f_S)$. ∎

CV stability gives necessary and sufficient conditions for distribution-dependent PAC learning. The bounds of Theorem 5.2 are analogous to those obtained using annealed entropy [10]. We conjecture that "CV theory" is to the distribution-dependent setting what VC theory is to the distribution-free setting.

**Corollary 5.4** *Let $\mathcal{H}$ be a space of $\pm 1$-classifiers. The following are equivalent:*

1. *There is a constant $K$ such that, for any distribution $\Delta$ on $X$, and any $h_0 \in \mathcal{H}$, ERM over $\mathcal{H}$ is $(0, e^{-Km})$-CV-stable (or, equivalently, $(0, e^{-Km})$-training-stable) with respect to the distribution on $\mathcal{Z}$ generated by $\Delta$ and $h_0$.*

2. $\mathrm{VCdim}(\mathcal{H}) < \infty$.

## 6 Examples of stable and unstable learners

We provide examples for a number of learning algorithms that may be analyzed within the framework of algorithmic stability.

**Example 6.1** *ERM over a finite $\mathcal{H}$*: If there exists a unique classifier $h_* \in \mathcal{H}$ which minimizes true error, then $\mathcal{A}$ is weakly $(0,\delta)$-hypothesis-stable for $\delta = \exp(-\Omega(m))$. Note that in general, such an $\mathcal{A}$ is *not* uniformly hypothesis stable.

**Example 6.2** *Regularization*: Bousquet and Elisseeff [1] prove that regularization is uniformly $\beta$-hypothesis-stable for $\beta = O(1/m)$.

**Example 6.3** *AdaBoost*: It is possible to show [7] that AdaBoost [4] is stability preserving: if the weak learner is uniformly hypothesis stable, then the strong learner is weakly hypothesis stable.

**Example 6.4** *Learning Finite Languages*: The hypothesis class $\mathcal{H}$ is consists of all finite languages over $\Sigma^*$ (where $\Sigma$ is a finite alphabet). Examples are drawn according to some distribution $D$ on $\Sigma^*$ and labeled according to a finite target language $L \subset \Sigma^*$, such that instance $x$ is labeled 1 if $x \in L$ and 0 if $x \notin L$.

Consider the learning algorithm $\mathcal{A}$ which, given $S$, returns the language $L_S = \{x : (x,1) \in S\}$. We ignore negative examples, and return the smallest language containing all positive examples. If $p = \min_{x \in L} \Pr(x)$, then $\mathcal{A}$ has weak hypothesis stability $\left(0, (1-p)^{m-O(\log m)}\right)$.

**Example 6.5** *ERM when* $\mathrm{VCdim}(\mathcal{H}) < \infty$: Kearns and Ron [5] prove that $\mathcal{A}$ is weakly error stable. In the PAC setting, $\mathcal{A}$ is training stable by Corollary 5.4.

**Example 6.6** *Maximum Margin Hyperplanes*: Suppose we are given a training set $\{(\mathbf{x}_i, y_i)\}$, with $\mathbf{x}_i \in \mathbb{R}^k$. The *maximum margin* hyperplane separating the data is the hyperplane $\mathbf{w} \cdot \mathbf{x} + b$ which minimizes $\mathbf{w} \cdot \mathbf{w}$, subject to $y_i(\mathbf{w} \cdot \mathbf{x}_i + b) \geq 1$ for every $i$.

Consider the action of the maximum margin algorithm on a training set $S$. There will be some subset of *support vectors* in $S$ for which $y_i(\mathbf{w} \cdot \mathbf{x}_i + b) = 1$. Suppose we start with $m+1$ points, and let $T$ be the set of support points. We now choose $z$ and $z_i$ from among our $m+1$ points, forming $S$ and $S'$. We will have $f_S = f_{S'}$ unless $z$ or $z_i$ lies inside $T$. It is possible to show:

**Theorem 6.7** *The maximum margin algorithm of Example 6.6 is weakly $(0,\delta)$-hypothesis-stable, where*

$$\delta = \frac{2\mathbf{E}_{S\sim D^{m+1}}(|T|)}{m+1},$$

*and $|T|$ is the number of support points among a training set of size $m + 1$.*

**Remark 6.8** Theorem 6.7 is reminiscent of a result of Vapnik [10, Theorem 10.5]: the expected error rate of support vector machines is at most $\frac{E(|T|)}{m-1}$.

## 7 Open questions

An important set of questions concern the relationship between the stability theory and the VC theory. For example, we show in the PAC setting that training stability is necessary and sufficient for learnability. In the VC framework, annealed entropy [10] can be used to obtain similar conditions. What is the relationship between training stability and annealed entropy? Is ERM over a space with low annealed entropy (i.e., $\lim_{m\to\infty} \mathcal{H}^{\mathcal{H}}_{\text{ann}}(m)/m = 0$) necessarily training stable? Can we prove a direct connection between training stability and annealed entropy, without connecting each to uniform convergence bounds?

A second class of questions relate to the application of the stability framework for the analysis of learning algorithms. For example, are algorithms like MDL, versions of boosting, decision trees, nearest neighbors, Bayesian learners stable in any sense?

A third class of questions relate to the possibility of *stability boosting* ensemble procedures. So far, boosting protocols like AdaBoost have focused on *error boosting*. Such algorithms provably boost training error and the implications for generalization error are unclear. In contrast, stability boosting would provably boost generalization error while potentially trading off training error.

## References

[1] O. Bousquet and A. Elisseeff. Stability and generalization. *Journal of Machine Learning Research*, 2:499–526, 2002.

[2] L. Breiman. Heuristics of instability and stabilization in model selection. *Annals of Statistics*, 24(6):2350–2383, 1996.

[3] L. Devroye and T. Wagner. Distribution-free performance bounds for potential function rules. *IEEE Trans. Inf. Thy.*, 25(5):601–604, 1979.

[4] Y. Freund and R. Schapire. A decision theoretic generalization of on-line learning and an application to boosting. *JCSS*, 55(1):119–139, 1997.

[5] M. Kearns and D. Ron. Algorithmic stability and sanity-check bounds for leave-one-out cross-validation. *Neural Computation*, 11:1427–1453, 1999.

[6] S. Kutin. Extensions to McDiarmid's inequality when differences are bounded with high probability. Technical report, Department of Computer Science, The University of Chicago, 2002. In preparation.

[7] S. Kutin and P. Niyogi. The interaction of stability and weakness in AdaBoost. Technical Report TR-2001-30, Department of Computer Science, The University of Chicago, 2001.

[8] S. Kutin and P. Niyogi. Almost-everywhere algorithmic stability and generalization error. Technical Report TR-2002-03, Department of Computer Science, The University of Chicago, 2002.

[9] C. McDiarmid. On the method of bounded differences. In *Surveys in combinatorics, 1989*, pages 148–188. Cambridge Univ. Press, 1989.

[10] V. Vapnik. *Statistical learning theory*. John Wiley & Sons Inc., New York, 1998.